\documentclass{article}

\usepackage[preprint]{icml2026/icml2026}

\usepackage{microtype}
\usepackage{graphicx}
\usepackage{booktabs}
\usepackage{amsmath,amssymb,amsfonts}
\usepackage{mathtools}
\usepackage{bm}
\usepackage{hyperref}
\usepackage{url}
\usepackage{natbib}
\usepackage{algorithm}
\usepackage{algorithmic}
\usepackage{caption}
\usepackage{placeins}
\usepackage{balance}

\DeclareMathOperator{\IoU}{IoU}
\DeclareMathOperator{\Len}{Len}
\DeclareMathOperator{\StopGrad}{stopgrad}
\newcommand{\stopgrad}[1]{\StopGrad\!\left(#1\right)}

\icmltitlerunning{Moondream Segmentation}
\makeatletter
\renewcommand{\Notice@String}{}
\makeatother

\begin{document}
\raggedbottom
\twocolumn[
  \icmltitle{Moondream Segmentation: From Words to Masks}

  \begin{icmlauthorlist}
    \icmlauthor{Ethan Reid}{m87}
  \end{icmlauthorlist}

  \icmlaffiliation{m87}{M87 Labs, San Francisco, CA, USA}
  \icmlcorrespondingauthor{Ethan Reid}{ethan@m87.ai, @EthanReidMorro}

  \vskip 0.3in
]

\printAffiliationsAndNotice{}  

\begin{abstract}
We present Moondream Segmentation, a referring image segmentation extension of Moondream~3, a vision-language model. Given an image and a referring expression, the model autoregressively decodes a vector path and iteratively refines the rasterized mask into a final detailed mask. We introduce a reinforcement learning stage that resolves ambiguity in the supervised signal by directly optimizing mask quality. Rollouts from this stage produce coarse-to-ground-truth targets for the refiner. To mitigate evaluation noise from polygon annotations, we release RefCOCO-M, a cleaned RefCOCO validation split with boundary-accurate masks. Moondream Segmentation achieves a cIoU of 80.2\% on RefCOCO (val) and 62.6\% mIoU on LVIS (val).
\end{abstract}

\section{Introduction}
Pixel-accurate segmentation is a core visual primitive for interactive and high-fidelity applications such as image editing and compositing. While modern vision-language models (VLMs) can localize and describe visual content, producing masks with precise boundaries, thin structures, and clean interiors from natural language remains challenging.

Referring image segmentation (RIS)~\citep{hu2016segmentation} takes an image and a referring expression (e.g., ``the car on the left'') and predicts a binary mask for the described region. RIS couples two failure modes: semantic grounding (identifying which instance the expression refers to) and boundary recovery (tracing exact contours under occlusion and fine detail). In practice, many systems either rely on multi-stage orchestration, where a language model produces prompts for a separate segmenter, as in SAM~3 Agent~\citep{carion2025sam3segmentconcepts}, or decode dense masks directly.

\begin{figure*}[t]
  \centering
  \includegraphics[width=\textwidth]{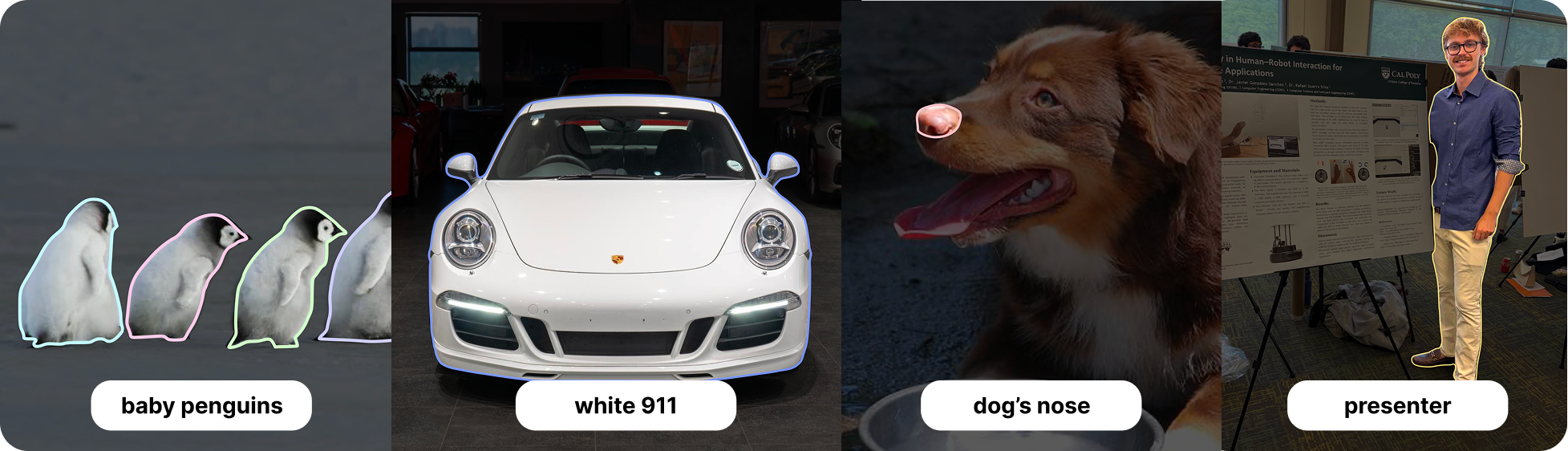}
  \caption{Example masks produced by Moondream Segmentation. Prompts are shown in white boxes.}
  \label{fig:example_masks}
\end{figure*}

\begin{figure*}[t]
  \centering
  \includegraphics[width=0.95\textwidth]{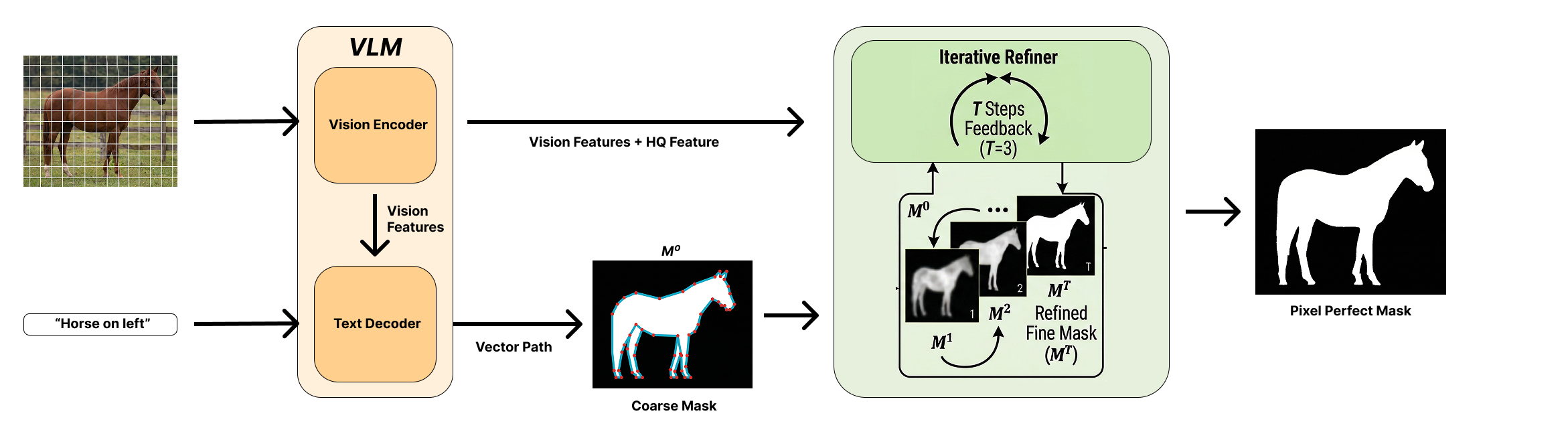}
  \caption{High-level overview of Moondream Segmentation. The VLM decodes a vector path from the image and prompt, which is rasterized into a coarse mask. An iterative refiner conditioned on frozen vision features produces the final mask.}
  \label{fig:overview}
\end{figure*}

Moondream Segmentation produces masks in two stages (Figure~\ref{fig:overview}). First, we autoregressively decode a vector path in an SVG-style syntax conditioned on the image and referring expression using Moondream~3~\citep{korrapati2025moondream}. Second, we rasterize the vector path into a coarse mask and iteratively refine the rasterization into a final detailed mask. Representative predictions on diverse scenes are shown in Figure~\ref{fig:example_masks}.

Supervising vector paths is inherently ambiguous: many different paths can rasterize to nearly identical masks. We address this with a reinforcement learning (RL) stage that directly optimizes mask overlap after rasterization. Rollouts from this stage produce intermediate coarse masks, which we reuse as coarse-to-ground-truth targets for training the refiner under the same iterative interface used at inference.

As models approach pixel-level accuracy, evaluation is increasingly limited by annotation fidelity. We evaluate on RefCOCO (UNC) and RefCOCO+ (UNC)~\citep{yu2016modeling} and RefCOCOg (Google)~\citep{mao2016generation}; to mitigate evaluation noise from polygon annotations, we release RefCOCO-M, a cleaned RefCOCO validation split with boundary-accurate masks and a small set of filtered expressions.

\noindent\textbf{Contributions.}
\begin{itemize}
  \item \textbf{Vector paths for RIS:} an extension of Moondream~3 that decodes a vector path as a compact intermediate representation and refines its rasterization into a detailed pixel mask.
  \item \textbf{RL for ambiguous supervision:} an RL stage that optimizes mask quality after rasterization and produces rollout-derived coarse masks used to train a refiner.
  \item \textbf{Boundary-accurate evaluation:} RefCOCO-M, a cleaned RefCOCO validation split with refined masks that reduces evaluation noise from polygon annotations.
\end{itemize}

\section{Related Work}

\noindent\textbf{Language-guided segmentation.} RIS benchmarks such as RefCOCO couple language grounding with mask annotations. Universal segmentation architectures (e.g., Mask2Former~\citep{cheng2022mask2former}) provide flexible transformer-based mask decoding across segmentation tasks. LISA~\citep{lai2024lisa} extends a large language model with a segmentation token that triggers a decoder to produce masks from referring and reasoning queries. Our goal is to expose a segmentation interface for an autoregressive VLM, while keeping boundary recovery in a dedicated refiner.

\noindent\textbf{Regions as sequences.} Several works cast localization and shape prediction as structured generation. Pix2Seq~\citep{chen2022pix2seq} models boxes as discretized coordinate tokens, while polygon-based methods such as Polygon-RNN++~\citep{acuna2018polygonrnnpp} and PolyFormer~\citep{liu2023polyformer} predict polygon vertices sequentially. More recently, VLMs have explored representing regions using special token sequences, e.g., discretized location tokens for detection and dedicated segmentation tokens for mask outputs~\citep{beyer2024paligemma}. We instead generate an explicit vector path whose rasterization defines a coarse region, then optimize the path with an overlap-based RL objective.

\noindent\textbf{Promptable segmentation and refiners.} Segment Anything (SAM) introduced a promptable segmentation interface with multiple mask hypotheses and a learned mask-quality predictor~\citep{kirillov2023sam}. We reuse this mask-token and quality-head pattern in our refiner, but the prompt is a model-generated coarse mask obtained by rasterizing the decoded vector path.

\section{Moondream~3 Backbone}
We build on Moondream~3, a 2B active / 9B total parameter mixture-of-experts (MoE) vision-language model~\citep{shazeer2017moe} that pairs a Vision Transformer (ViT) vision encoder~\citep{dosovitskiy2021vit} with a decoder-only language model. The vision encoder is initialized from SigLIP~2 weights~\citep{tschannen2025siglip2}, and the text backbone has 24 layers at width $D=2048$; from layer 4 onward, its feed-forward blocks are MoE with 64 experts and top-8 routing. In this paper we only describe the components needed for segmentation: the vision tokenization scheme (which defines the $27\times 27$ grid used by our refiner) and the RegionModel, a region-tokenizer interface for binned spatial values (used for points and boxes).

\subsection{Vision Encoder}
Moondream~3 converts an image into a fixed-length set of patch tokens by running a ViT over $378\times 378$ crops with patch size $14$, producing a $27\times 27$ grid (729 tokens) with width $1152$. To support high-resolution images, it uses one global crop and up to $12$ overlapping local crops (overlap margin $4$ patches). Local token grids are stitched with overlap-aware reconstruction and pooled back to $27\times 27$; the global and pooled-local features are concatenated (2304-dim) and projected to the text width $D=2048$ before entering the text transformer.

\subsection{RegionModel}
Moondream represents spatial values using a small region tokenizer implemented with special tokens \texttt{<|coord|>} and \texttt{<|size|>}. A point is represented as \texttt{<|coord|><|coord|>} (center $(c_x,c_y)$). A box is represented as \texttt{<|coord|><|coord|><|size|>} (center $(c_x,c_y)$ and size $(w,h)$).

At generation time, when the model emits a coordinate or size token, a small RegionModel decodes a 1024-dim distribution from the current hidden state, maps the selected bin back to a real value, and immediately re-embeds it for the next decoding step. This enables structured spatial outputs without introducing a large numeric vocabulary. Appendix~\ref{app:regionmodel} provides the RegionModel encoding/decoding equations and the decode-then-re-embed procedure.

\section{Moondream Segmentation}
\subsection{Task}
Given an image $I\in\mathbb{R}^{H\times W\times 3}$ and a referring expression $x$ (text and/or a spatial reference), the goal is to predict a binary mask $M\in\{0,1\}^{H\times W}$ for the region described by $x$. We evaluate using intersection-over-union:
\begin{equation}
\IoU(\hat{M}, M)=\frac{|\hat{M}\cap M|}{|\hat{M}\cup M|},
\end{equation}
where $\hat{M}$ is the predicted mask. We view masks as sets of foreground pixels, where $|\cdot|$ counts pixels.

\subsection{Output Representation}
We represent the coarse mask as a predicted bounding box $b$ and a set of closed vector paths $p$ represented as a command-and-coordinate sequence, with commands such as \texttt{M}, \texttt{L}, \texttt{C}, and \texttt{Z}. The box is produced using Moondream's RegionModel interface (special \texttt{<|coord|>} and \texttt{<|size|>} tokens; Appendix~\ref{app:regionmodel}). The path is then emitted autoregressively as text tokens conditioned on $(I,x,b)$. A deterministic rasterizer maps $(b,p)$ into an initial coarse mask. We rasterize at the native image resolution and resize to the refiner resolution:
\begin{equation}
\begin{aligned}
\tilde{M}^{(0)}_{\text{nat}} &= \mathrm{Rasterize}(b,p; H, W)\in[0,1]^{H\times W},\\
\tilde{M}^{(0)} &= \mathrm{Resize}(\tilde{M}^{(0)}_{\text{nat}}; H_0, W_0)\in[0,1]^{H_0\times W_0}.
\end{aligned}
\end{equation}
We use a fixed refiner resolution $H_0=W_0=378$ in all experiments.

\paragraph{Decoding constraints.}
We constrain decoding to ensure syntactic validity and stable geometry. Concretely, we enforce a maximum path complexity $\Len(p)\le L_{\max}$ and reject invalid paths; Appendix~\ref{app:svg} provides the full token sequence format and vector path token grammar.

\subsection{Mask Refiner}
The mask refiner maps a coarse raster mask to a boundary-accurate prediction conditioned on frozen Moondream vision features. Architecturally, it separates \emph{dense evidence} (image tokens) from \emph{mask hypotheses} (a small set of learned output tokens) and updates both via two-way attention. This mirrors the ``mask token + quality head'' pattern introduced by SAM, but our refiner is used iteratively with a coarse mask as a dense prompt.

\paragraph{Inputs and outputs.}
The refiner operates on features from the global crop only, using the 1152-dim vision encoder output before the multi-crop concatenation and projection to $D$. At refinement step $t$, inputs are final-layer patch tokens $F_{\text{final}}\in\mathbb{R}^{B\times 729\times 1152}$, early-layer patch tokens $F_{\text{early}}\in\mathbb{R}^{B\times 729\times 1152}$, and the current mask estimate $\tilde{M}^{(t)}\in[0,1]^{B\times 1\times H_0\times W_0}$. We obtain $F_{\text{early}}$ by exposing an intermediate token grid from an earlier ViT block (block 8) in addition to the final tokens, providing higher-frequency cues for refinement. The refiner outputs $K$ mask logits $L^{(t)}\in\mathbb{R}^{B\times K\times H_0\times W_0}$ and mask-quality scores $q^{(t)}\in\mathbb{R}^{B\times K}$.

\paragraph{Grid features and HQ fusion.}
We reshape patch tokens to a $27\times 27$ grid (729 tokens) and project the 1152-dim Moondream patch embeddings to a refiner width $d_r=256$. This matches the vision encoder's patch size of $14$ at our fixed refiner resolution $H_0=W_0=378$ (since $27\cdot 14=378$). Let $F_{\text{fused}}\in\mathbb{R}^{B\times d_r\times 27\times 27}$ denote fused features obtained by combining early and final vision features with lightweight convolutions, following the high-quality feature fusion design as seen in HQ-SAM~\citep{ke2023segmenthighquality}.

\paragraph{Dense mask prompting.}
A convolutional mask encoder downsamples $\tilde{M}^{(t)}$ to $27\times 27$ and lifts it to $d_r$ channels, producing $E_{\text{mask}}\in\mathbb{R}^{B\times d_r\times 27\times 27}$. We prompt the refiner by additive fusion:
\begin{equation}
X_{\text{grid}} = F_{\text{fused}} + E_{\text{mask}}.
\end{equation}
We then flatten $X_{\text{grid}}$ into image tokens $X\in\mathbb{R}^{B\times 729\times d_r}$.

\paragraph{Two-way token interaction.}
We maintain a set of learned output tokens $Z\in\mathbb{R}^{B\times (K+1)\times d_r}$ consisting of $K$ mask tokens and one quality token. Let $X\in\mathbb{R}^{B\times N\times d_r}$ be the image tokens ($N=729$). A two-way transformer alternates self-attention on $Z$, cross-attention $Z\rightarrow X$, an MLP on $Z$, and cross-attention $X\rightarrow Z$, updating both token sets. Abstractly, each block can be written as
\begin{equation}
(Z',X')=\mathrm{TwoWayBlock}(Z,X),
\end{equation}
where attention uses learned positional embeddings on $Q/K$ (SAM-style).

\paragraph{Hypernetwork decoding and mask-quality head.}
We upsample image tokens to a high-resolution feature map $U\in\mathbb{R}^{B\times C\times 216\times 216}$ (three $2\times$ upsampling stages from the $27\times 27$ grid). Each mask token $Z_m$ is first projected and refined by high-resolution cross-attention, yielding a per-mask summary vector $s_m\in\mathbb{R}^{C}$. A small per-mask MLP (hypernetwork) maps $s_m$ to channel weights $w_m\in\mathbb{R}^{C}$, and mask logits are computed by a channel-weighted sum
\begin{equation}
L^{(t)}_m(i,j)=\sum_{c=1}^{C} w_{m,c}\,U_{c}(i,j),
\end{equation}
followed by resizing to $(H_0,W_0)$. The quality token is mapped by an MLP to per-mask scores $q^{(t)}$, which we use to select the hypothesis at each iteration, following SAM.

\paragraph{Iterative refinement.}
Refinement is applied for $T=5$ iterations at both training and inference. At each iteration we select the highest-scoring mask and feed it back as the next estimate. We obtain the final prediction at the native image resolution by resizing the last refined mask: $\hat{M}=\mathrm{Resize}(\tilde{M}^{(T)}; H, W)$. During training, we apply a stop-gradient operator to the updated mask between iterations; inference does not require stop-gradient. Pseudocode for the inference loop is provided in Appendix~\ref{app:refiner_inference}.
We evaluate refiner quality with ablations in Table~\ref{tab:ablations}.

\section{Training}
We train the system in three stages: supervised fine-tuning (SFT) for vector path generation, RL to align the path decoder with mask quality, and supervised refiner training using rollout-generated coarse masks.

\subsection{Supervised Training}

\paragraph{Data pipeline.}
We construct a training set of over 10M samples from internet images. For each image, an ensemble of VLMs produces tags, referring expressions, descriptions, and grounded bounding boxes. We verify the text--box pairs with Moondream. For a given image and text prompt, we discard samples where Moondream predicts more than one box or the predicted box has less than 90\% IoU with the annotation. Surviving image--box pairs are fed to a segmentation model that produces and iteratively refines a mask for each box. We apply a second filter, discarding samples where the mask-derived bounding box falls below 92\% IoU with the Moondream box. Train splits of public datasets such as RefCOCO/+/g and LVIS are not used directly; we run their images through the same pipeline, optionally retaining original referring expressions subject to the same filtering. Each final sample pairs an image and referring expression with a mask.

\begin{figure*}[t]
  \centering
  \includegraphics[width=0.95\textwidth]{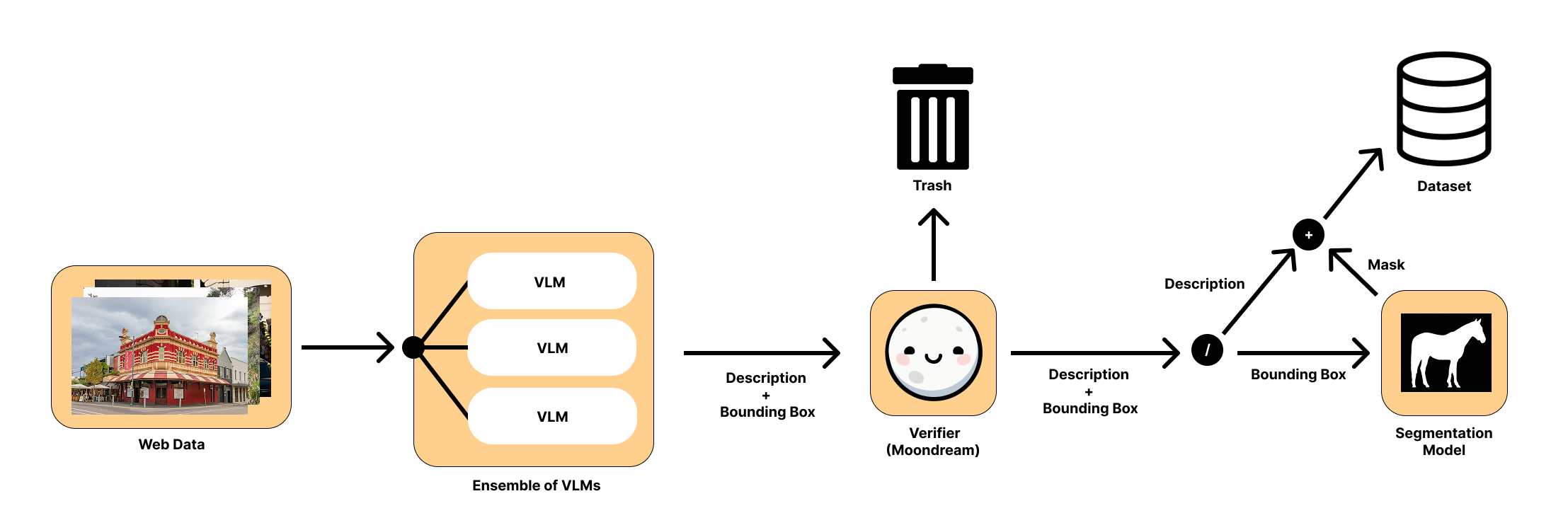}
  \caption{Training data generation pipeline. Web images are labeled by an ensemble of VLMs with text annotations and bounding boxes, verified by Moondream, filtered for consistency and accuracy, and passed to a segmentation model to propose masks. Surviving image--text--box--mask tuples are added to the final dataset; rejected samples are discarded.}
  \label{fig:data_pipeline}
\end{figure*}

\paragraph{Training objective.}
We train the path decoder with SFT on this corpus. Each mask is converted to a target vector path token sequence $p_{1:L}$. We train with teacher forcing and minimize the standard token-level cross-entropy objective
\begin{equation}
\mathcal{L}_{\text{SFT}} = -\sum_{k=1}^{L} \log \pi_{\theta}(p_k\mid p_{<k}, I, x),
\end{equation}
without any additional format-specific penalties.

\subsection{RL for Path Optimization}
\paragraph{Why next-token supervision is misaligned.}
SFT treats vector path generation as a standard next-token prediction problem. However, vector paths admit many equivalent tokenizations: different starting points, equivalent segment decompositions, or alternative control points can rasterize to essentially the same mask. This induces a one-to-many mapping from an underlying region to token sequences.

In practice, this ambiguity is most damaging at the beginning of decoding. The ``first few tokens'' (e.g., initial \texttt{M} command and early coordinates) are effectively arbitrary choices among many equivalent parameterizations; as a result, the model is penalized for producing a valid but differently parameterized path. This creates high-variance gradients and can destabilize learning, even when the downstream rasterized mask would be correct.

\paragraph{RL aligns learning to what matters.}
To resolve this mismatch, we fine-tune the path policy $\pi_{\theta}$ with RL so that \emph{any} valid sequence that yields a high-quality mask is rewarded. Rather than matching an arbitrary reference tokenization, the objective is defined on the rasterized output and validity constraints:
\begin{equation}
\max_{\theta}\; \mathbb{E}_{(b,p)\sim \pi_{\theta}(\cdot\mid I,x)}\big[ R(\mathrm{Rasterize}(b,p), M)\big].
\end{equation}
We optimize this objective using Group Relative Policy Optimization (GRPO)~\citep{shao2024deepseekmath}.

As a result, RL can improve both correctness (IoU/boundary metrics) and generation stability (valid path rate and reduced sensitivity to early-token choices).

\subsubsection{Piecewise Reward}
\label{sec:piecewise_reward}
We use a piecewise reward that changes with model competence to maximize learning signal while avoiding reward noise early in training.

\paragraph{Group reward with invalid-trace filtering.}
We sample a group $\mathcal{G}$ of rollouts for each $(I,x)$ and use rollout-excluded (unbiased) reward selection. Some rollouts may be invalid paths (e.g., wrong arity for a \texttt{C} command). We assign these rollouts reward $0$ and exclude them from the group-average used to choose the reward regime. Let $\mathcal{G}_{\text{valid}}\subseteq\mathcal{G}$ be valid rollouts, and let $\bar{\IoU}_{\text{box}}$ and $\bar{\mathrm{Tv}}_{\alpha,\beta}$ denote means over $\mathcal{G}_{\text{valid}}$.

\paragraph{Bounding-box IoU.}
Let $b^*$ denote the tight bounding box of the target mask $M$.
For boxes $b$ and $b^*$ with areas $|\cdot|$, we define
\begin{equation}
\IoU_{\text{box}}(b,b^*) = \frac{|b\cap b^*|}{|b\cup b^*|}.
\end{equation}

\paragraph{Tversky for coverage.}
For masks, define $\mathrm{TP}=|\tilde{M}\cap M|$, $\mathrm{FN}=|M\setminus \tilde{M}|$, and $\mathrm{FP}=|\tilde{M}\setminus M|$. The Tversky index~\citep{tversky1977features} is
\begin{equation}
\mathrm{Tv}_{\alpha,\beta}(\tilde{M}, M)=\frac{\mathrm{TP}}{\mathrm{TP}+\alpha\,\mathrm{FP}+\beta\,\mathrm{FN}}.
\end{equation}
We set $\beta>\alpha$, which penalizes false negatives more than false positives, biasing the model toward \emph{coverage} (reducing missed parts), which we found important early in training.

\paragraph{Boundary IoU for edge tightening.}
Following Boundary IoU~\citep{cheng2021boundaryiou}, we define a boundary band operator that keeps only pixels inside a mask and within distance $\varepsilon$ of that mask's boundary. Let $\mathrm{dist}_{A}(i,j)$ denote the distance from pixel $(i,j)$ to the boundary of a binary mask $A$ measured from inside $A$. We define
\begin{equation}
\big(\mathcal{B}_{\varepsilon}(A)\big)_{i,j} = A_{i,j} \cdot \mathbb{I}\big[\mathrm{dist}_{A}(i,j)\le \varepsilon\big].
\end{equation}
We then define boundary IoU as the IoU between the boundary bands of the prediction and target:
\begin{equation}
\label{eq:boundary_iou}
\IoU_{\partial}(\tilde{M},M)=\IoU\big(\mathcal{B}_{\varepsilon}(\tilde{M}),\; \mathcal{B}_{\varepsilon}(M)\big),
\end{equation}
which concentrates reward on edges and thin structures once coarse coverage is achieved. For RL training, we set $\varepsilon$ to 5\% of the image diagonal.

\paragraph{Unbiased piecewise reward.}
For each valid rollout $g$ producing a box $b_g$ and mask $\tilde{M}_g$, we define the per-rollout reward
\begin{equation}
 r_g =
 \begin{cases}
  \IoU_{\text{box}}(b_g,b^*) & \text{if } \bar{\IoU}_{\text{box}} < \tau_{\text{box}},\\
  \mathrm{Tv}_{\alpha,\beta}(\tilde{M}_g,M) &
  \begin{array}[t]{@{}l@{}}
  \text{if } \bar{\IoU}_{\text{box}} \ge \tau_{\text{box}}\\
  \text{and } \bar{\mathrm{Tv}}_{\alpha,\beta} < \tau_{\text{mask}}
  \end{array}
  ,\\
  \IoU_{\partial}(\tilde{M}_g,M) & \text{otherwise.}
 \end{cases}
\end{equation}
Invalid rollouts receive $r_g=0$ but are excluded from the group means $\bar{\IoU}_{\text{box}}$ and $\bar{\mathrm{Tv}}_{\alpha,\beta}$.

This curriculum aims to accelerate learning: box IoU teaches object discovery, Tversky improves coverage, and boundary IoU sharpens edges, producing high-quality coarse masks that become easy-to-optimize targets for the refiner.

\begin{table*}[t]
  \centering
  \footnotesize
  \setlength{\tabcolsep}{3.5pt}
  \caption{RIS benchmark comparison (\%; higher is better). RefCOCO/+/g and RefCOCO-M report cIoU; the final RefCOCO-M column reports BIoU@0.05.}
  \label{tab:main_results}
  \begin{tabular}{@{}lrrrrrrrrrr@{}}
    \toprule
    & \multicolumn{3}{c}{RefCOCO} & \multicolumn{3}{c}{RefCOCO+} & \multicolumn{2}{c}{RefCOCOg} & \multicolumn{2}{c}{RefCOCO-M} \\
    \cmidrule(lr){2-4} \cmidrule(lr){5-7} \cmidrule(lr){8-9} \cmidrule(lr){10-11}
    Model & val & testA & testB & val & testA & testB & val & test & val & BIoU@0.05 \\
    \midrule
    LISA (LLaVA~7B) & 74.9 & 79.1 & 72.3 & 65.1 & 70.8 & 58.1 & 67.9 & 70.6 & 72.7 & 68.1 \\
    Text4Seg (InternVL2-8B) & 74.7 & 77.4 & 71.6 & 68.5 & 63.6 & 62.9 & 70.7 & 71.6 & 70.4 & 68.9 \\
    SimpleSeg (Qwen2.5-VL) & \textbf{80.9} & 77.8 & 75.2 & 72.4 & 77.3 & \textbf{66.1} & 73.3 & \textbf{74.1} & 85.2 & 68.9 \\
    Gemini~2.5 Flash & 63.7 & 64.8 & 65.6 & 52.8 & 57.3 & 50.8 & 56.2 & 57.4 & 70.4 & 67.5 \\
    SAM~3 Agent (Gemini~2.5 Pro) & 75.5 & 77.6 & 71.0 & 67.3 & 71.1 & 63.4 & 73.4 & 74.0 & 86.7 & 81.5 \\
    SAM~3 & 54.9 & 59.9 & 52.4 & 43.0 & 49.7 & 36.7 & 55.4 & 54.6 & 63.0 & 64.0 \\
    Moondream Seg. & 80.2 & \textbf{80.3} & \textbf{75.9} & \textbf{72.5} & \textbf{77.8} & 65.1 & \textbf{73.7} & 73.9 & \textbf{87.6} & \textbf{85.4} \\
    \bottomrule
  \end{tabular}
\end{table*}

\subsection{Refiner Training}
We train the refiner on an offline dataset of rollout snapshots, denoted $\mathcal{D}$, where each example provides an image $I$, an initial coarse mask $\tilde{M}^{(0)}$ obtained by rasterizing a model-generated vector path, and a target mask $M$. We unroll refinement for $T=5$ steps, matching the iterative test-time interface. Let $E$ denote the frozen vision encoder (used only to compute $(F_{\text{early}},F_{\text{final}})$), and let $f_{\phi}$ denote the refiner. We write $\sigma$ for the elementwise sigmoid. We also define a per-step loss $\mathcal{L}_{\text{ref-step}}(L^{(t)},q^{(t)},M)$ as the bracketed term in Eq.~(\ref{eq:ref_loss}).

\begin{algorithm}[t]
\caption{Refiner training on rollout-derived coarse masks (T=5)}
\label{alg:refiner_train}
\begin{algorithmic}[1]
\STATE \textbf{Input:} rollout dataset $\mathcal{D}$ of $(I,\tilde{M}^{(0)},M)$; frozen encoder $E$; refiner $f_{\phi}$; steps $T=5$
\FOR{each minibatch $(I,\tilde{M}^{(0)},M)\sim\mathcal{D}$}
  \STATE $(F_{\text{early}},F_{\text{final}})\leftarrow \stopgrad{E(I)}$
  \STATE $\tilde{M}\leftarrow \tilde{M}^{(0)}$; $\mathcal{L}\leftarrow 0$
  \FOR{$t=0$ to $T-1$}
    \STATE $(L^{(t)}, q^{(t)}) \leftarrow f_{\phi}(F_{\text{final}},F_{\text{early}},\tilde{M})$
    \STATE $\mathcal{L} \mathrel{+}= \mathcal{L}_{\text{ref-step}}(L^{(t)}, q^{(t)}, M)$ \COMMENT{Eq.~(\ref{eq:ref_loss}); losses apply to all $K$ masks}
    \STATE $m^* \leftarrow \arg\max_m q^{(t)}_m$
    \STATE $\tilde{M} \leftarrow \stopgrad{\sigma(L^{(t)}_{m^*})}$
  \ENDFOR
  \STATE Update $\phi$ on $\mathcal{L}/T$
\ENDFOR
\end{algorithmic}
\end{algorithm}

\paragraph{Feedback and stop-gradient.}
At each refinement step we select $m^*=\arg\max_m q^{(t)}_m$ and feed back $\tilde{M}^{(t+1)}=\stopgrad{\sigma(L^{(t)}_{m^*})}$ during training. This matches the iterative test-time interface while keeping gradients local to each refinement step.

\paragraph{Loss.}
We supervise every step against the same target mask $M$. In Algorithm~\ref{alg:refiner_train} we write $\mathcal{L}_{\text{ref-step}}(L^{(t)},q^{(t)},M)$ for the per-step refiner loss, defined as the bracketed term in Eq.~(\ref{eq:ref_loss}). Crucially, the segmentation loss is applied to \emph{all} $K$ mask hypotheses at each step (not just the selected hypothesis), which provides dense gradient signal while still training the model to select the best mask via the quality head. Concretely, we optimize
\begin{equation}
\label{eq:ref_loss}
\begin{aligned}
\mathcal{L}_{\text{ref}}
&=\frac{1}{T}\sum_{t=0}^{T-1}\Big[\;
\mathrm{SegLoss}(L^{(t)},M)\\
&\quad+\lambda_{\text{iou}}\,\|q^{(t)}-\hat{q}^{(t)}\|_2^2\\
&\quad+\lambda_{\partial}(s)\,\mathrm{BoundaryLoss}(\sigma(L^{(t)}),M)
\Big],
\end{aligned}
\end{equation}
where $\mathrm{SegLoss}$ combines binary cross-entropy (BCE) with a Dice loss term~\citep{milletari2016vnet} (averaged over the $K$ hypotheses), $\hat{q}^{(t)}=\stopgrad{\mathrm{SoftIoU}(\sigma(L^{(t)}),M)}$ is a differentiable soft-IoU target (Appendix~\ref{app:softiou}) used to supervise the quality head (computed without backprop through the target), and $\mathrm{BoundaryLoss}$ is a boundary-weighted BCE term (also averaged over hypotheses). We use a boundary-loss schedule $\lambda_{\partial}(s)$ that is $0$ early in training and is linearly warmed up to $1$ over a fixed step window; we elaborate the exact schedule and weighting in Appendix~\ref{app:impl}. The quality head is trained in the SAM style via regression to $\mathrm{SoftIoU}$ targets.

\section{Experiments}
\subsection{Datasets}
\paragraph{RefCOCO/+/g.}
We evaluate referring image segmentation on the RefCOCO (UNC), RefCOCO+ (UNC), and RefCOCOg (Google) evaluation splits, which pair COCO images~\citep{lin2014microsoft} with referring expressions and instance-level masks.

\begin{figure}[t]
  \centering
  \includegraphics[width=0.98\linewidth]{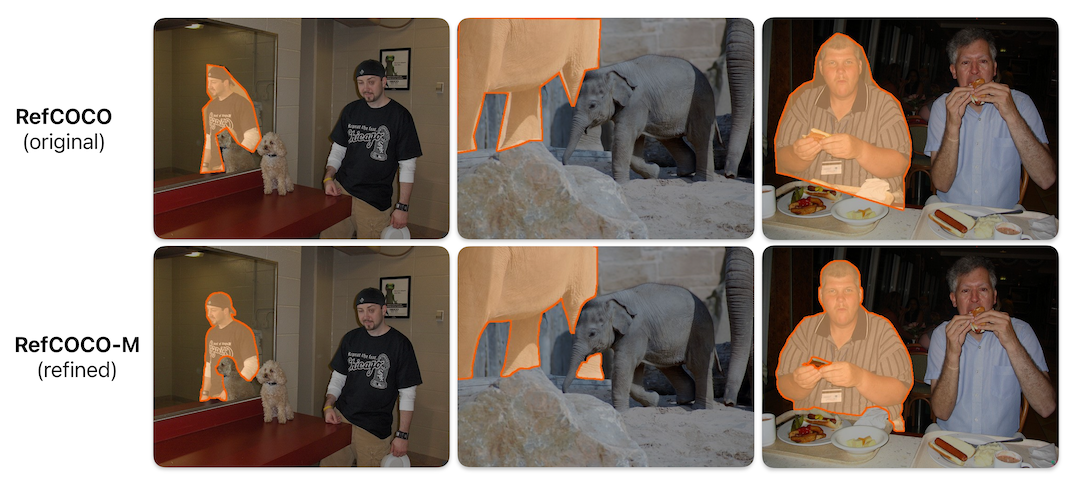}
  \caption{Original RefCOCO polygon masks (top) and RefCOCO-M refined masks (bottom). RefCOCO-M tightens boundaries and recovers fine structure that is often missing from the original annotations.}
  \label{fig:refcocom_compare}
\end{figure}

\paragraph{RefCOCO-M.}
RefCOCO's polygon-derived masks can be coarse at object boundaries and may miss thin structures, introducing evaluation noise as models approach pixel-level accuracy. We release RefCOCO-M\footnote{\url{https://huggingface.co/datasets/moondream/refcoco-m}}, a cleaned version of the RefCOCO (UNC) validation split. For each referred instance, we run a re-segmentation pipeline with an ensemble of models and retain only high-confidence masks, removing 47\% of instances due to unrecoverable mask quality. A separate model filters a further 0.5\% of samples (46 examples) for harmful language, including slurs, sexualized language, and degrading phrases (Figure~\ref{fig:filtered_samples}). The final dataset contains 1,190 images, 2,080 instance masks, and 5,598 referring expressions. For retained examples, the images and referring expressions are unchanged and only the masks differ (Figure~\ref{fig:refcocom_compare}).

For RefCOCO/+/g and RefCOCO-M, we report cIoU,
\begin{equation}
\label{eq:ciou}
\mathrm{cIoU}=\frac{\sum_n |\hat{M}_n \cap M_n|}{\sum_n |\hat{M}_n \cup M_n|},
\end{equation}
where the sums run over all $n$ evaluated samples. For RefCOCO-M, we additionally report BIoU@0.05, computed using the Boundary IoU in Eq.~(\ref{eq:boundary_iou}) with boundary width set to 5\% of the image diagonal.

\paragraph{LVIS.}
To measure instance segmentation performance on the LVIS validation split~\citep{gupta2019lvis}, we report mIoU. For each image, we prompt the model with every annotated category name and collect the predicted masks. Predicted and ground-truth masks are paired via Hungarian matching on IoU; any unpaired mask scores zero, penalizing both under- and over-prediction.

\subsection{Baselines}
We compare against six external baselines spanning promptable segmentation, direct VLM mask generation, tool-augmented pipelines, and recent open referring segmentation models: SAM~3, Gemini~2.5 Flash, SAM~3 Agent (Gemini~2.5 Pro), LISA (LLaVA~7B), Text4Seg (InternVL2-8B), and SimpleSeg (Qwen2.5-VL).

\paragraph{SAM 3.}
We evaluate SAM 3 as a promptable segmentation model with text prompts. It is important to note that SAM~3 is not presented as an RIS model, but as a Promptable Concept Segmentation model, where a prompt is a short noun phrase, an image exemplar, or a combination of both, and the model returns masks for all matching object instances. In our RefCOCO-family evaluation, this restricted language ability is the main bottleneck: the dominant failure mode is semantic grounding of referring expressions, rather than mask quality once the target region has been identified. For each referring expression, we pass the referring text directly as the SAM 3 text prompt and select the highest-scoring predicted mask.

\paragraph{Gemini 2.5 Flash.}
We follow the official Gemini API segmentation recipe~\citep{googledevai2026imageunderstanding}. We prompt the model to return a JSON list that includes a 2D bounding box and a base64-encoded segmentation mask cropped to that box. We resize the returned mask to the box size and paste it into the full image canvas for evaluation.

\paragraph{SAM 3 Agent (Gemini 2.5 Pro).}
We implement a SAM 3 Agent (Gemini 2.5 Pro) using the Gemini API~\citep{googledevai2026imageunderstanding} and the official SAM 3 repository~\citep{sam3repo}. The agent iteratively proposes short text phrases, calls SAM 3 for mask proposals, and selects the final mask to match the original referring expression.

\paragraph{LISA (LLaVA 7B).}
We evaluate the official LISA-7B-v1 release using the official LISA inference code path~\citep{lisarepo}. For each referring expression, we use the prompt template from the repository, ``\{expr\} Please output segmentation mask.''

\paragraph{Text4Seg (InternVL2-8B).}
We evaluate Text4Seg (InternVL2-8B) using the released Text4Seg inference stack~\citep{text4seg,text4segrepo}. The model uses InternVL2-8B with a SAM ViT-H refinement stage, and we use the prompt template from the repository, ``Please segment only the \{expr\} in the image.''

\paragraph{SimpleSeg (Qwen2.5-VL).}
We evaluate the released SimpleSeg-Qwen2.5-VL model using the official SimpleSeg repository guidance~\citep{simplesegrepo}. We use the prompt template from the repository, ``Output the polygon coordinates of \{expr\} in the image.'', and the slow processor path for best performance.

\subsection{Main Results}
\begin{table}[H]
  \centering
  \footnotesize
  \caption{LVIS (val) instance segmentation comparison (mIoU, \%; higher is better).}
  \label{tab:lvis_results}
  \begin{tabular}{lc}
    \toprule
    Model & LVIS \\
    \midrule
    Gemini~2.5~Flash & 45.5 \\
    SAM~3 Agent (Gemini~2.5 Pro) & 59.3 \\
    SAM~3 & \textbf{62.6} \\
    Moondream Seg. & \textbf{62.6} \\
    \bottomrule
  \end{tabular}
\end{table}

\paragraph{Quantitative results.}
Table~\ref{tab:main_results} shows that Moondream Segmentation outperforms SAM~3, Gemini~2.5 Flash, LISA, and Text4Seg on all RefCOCO variants, and is competitive with SimpleSeg, leading on RefCOCO testA and testB while trailing by 0.7 percentage points on RefCOCO val. Relative to SAM~3 Agent (Gemini~2.5 Pro), Moondream Segmentation scores higher on all reported RefCOCO-family splits except RefCOCOg test, while requiring no external tool orchestration. On RefCOCO-M, Moondream Segmentation achieves the strongest results, reaching 87.6 cIoU and 85.4 BIoU@0.05. Table~\ref{tab:lvis_results} shows that Moondream Segmentation and SAM~3 both reach 62.6 mIoU on LVIS, ahead of SAM~3 Agent (Gemini~2.5 Pro) at 59.3 and Gemini~2.5 Flash at 45.5. RefCOCO and LVIS validation samples are shown in Appendix Figures~\ref{fig:refcoco_qual} and~\ref{fig:lvis_qual}.

\begin{figure}[t]
  \centering
  \includegraphics[width=0.95\linewidth]{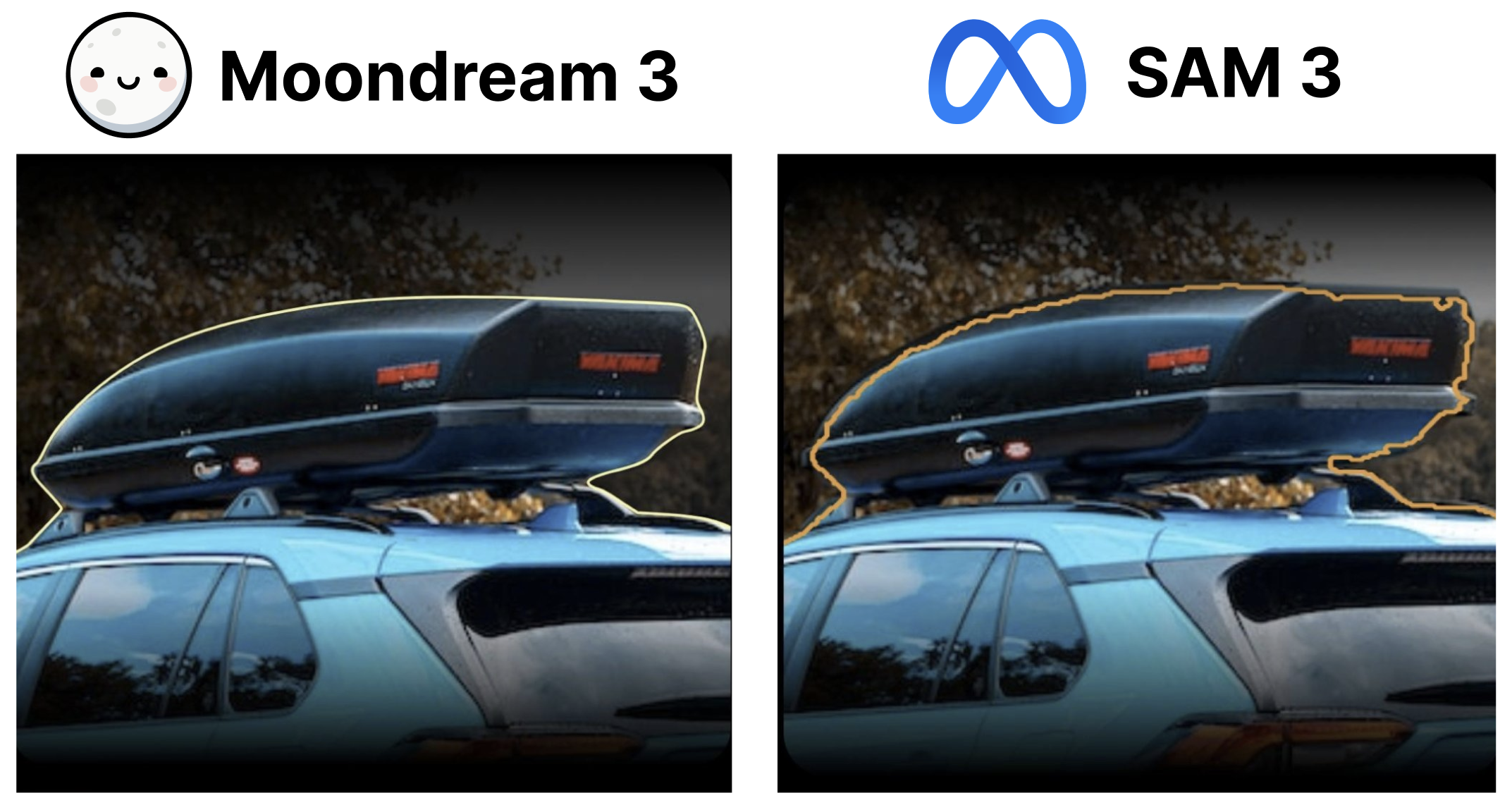}
  \caption{Boundary-focused qualitative comparison (prompt: \texttt{car}). Moondream masks are typically sharper at edges and better preserve fine structure than SAM~3.}
  \label{fig:sam3_boundary}
\end{figure}

\begin{figure}[t]
  \centering
  \includegraphics[width=0.95\linewidth]{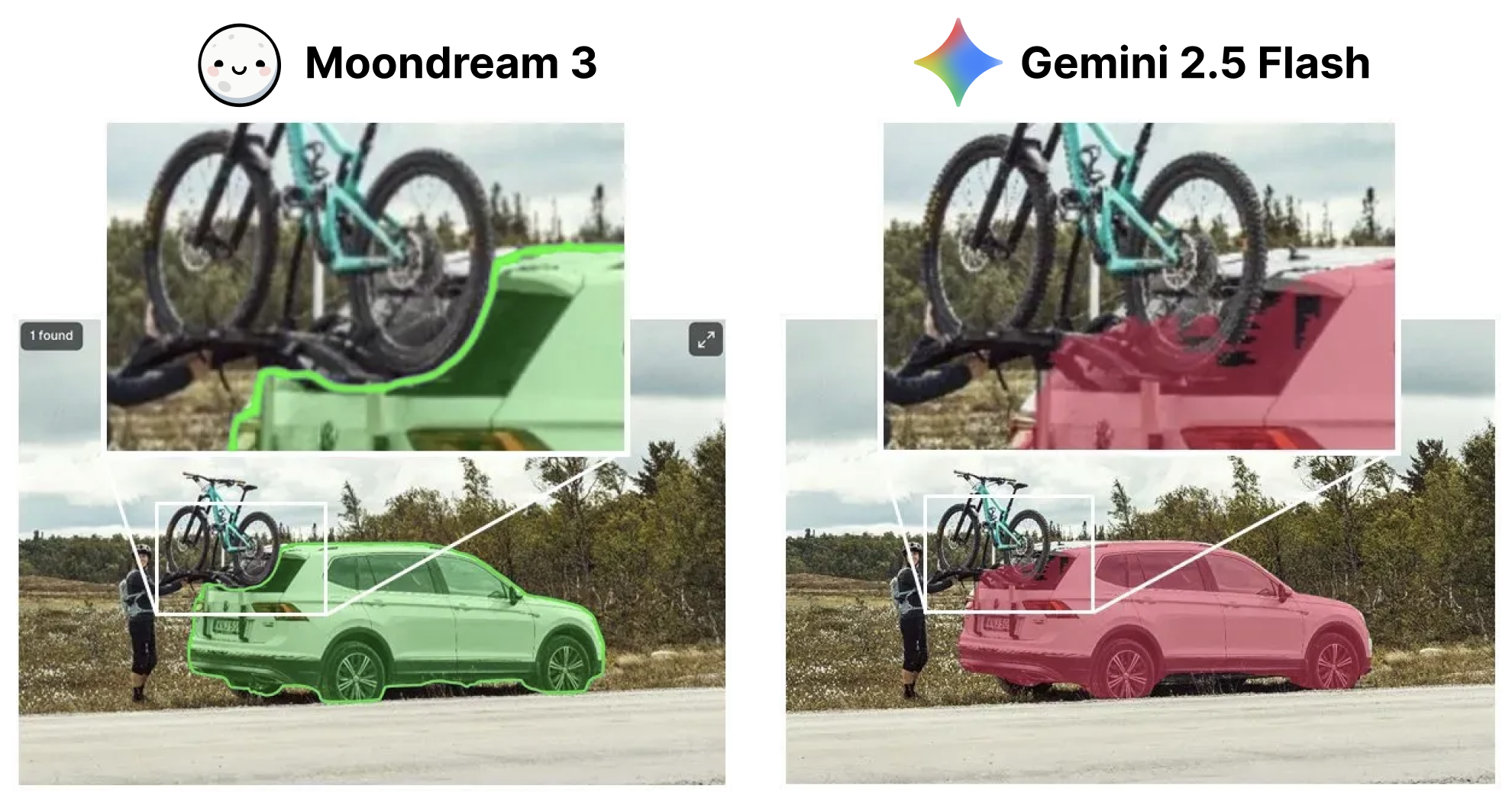}
  \caption{Qualitative comparison against Gemini~2.5 Flash (prompt: \texttt{car}). Gemini masks can contain small isolated false positives (salt-and-pepper noise) and boundary noise, while Moondream produces cleaner masks with sharper edges.}
  \label{fig:gemini_speckling}
\end{figure}

\paragraph{Boundary fidelity.}
Beyond aggregate metrics, boundary sharpness matters for high-fidelity downstream use, where small boundary errors are visually salient. Figures~\ref{fig:sam3_boundary} and~\ref{fig:gemini_speckling} compare Moondream Segmentation against SAM~3 and Gemini~2.5 Flash. SAM~3 often produces softer boundaries around fine structure, while our vector path decoding followed by refinement yields sharper edges that better align to visible contours. Gemini~2.5 Flash can contain small isolated false positives, a failure mode commonly described as ``salt-and-pepper'' artifacts in pixel-wise segmentation outputs, while our masks are typically cleaner and more spatially consistent.

\begin{figure}[t]
  \centering
  \includegraphics[width=0.95\linewidth]{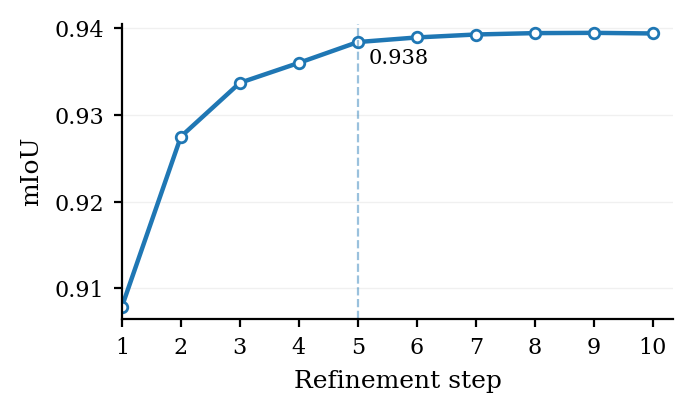}
  \caption{mIoU on a held-out set of RefCOCO-M rollout-derived coarse masks conditioned on the ground-truth bounding box, as a function of refinement steps.}
  \label{fig:refinement_steps}
\end{figure}

\subsection{Refiner Comparison}
We compare our refiner against SAM-based alternatives on RefCOCO-M using rollout-derived coarse masks conditioned on the ground-truth bounding box. This isolates refinement quality from semantic grounding errors by avoiding cases where the coarse mask does not correspond to the target instance. Each method takes the same rollout-derived coarse mask as input and runs for $T=5$ refinement steps. We report mIoU between the final refined mask and the ground truth. Note that the SAM-based refiners, SAM~2~\citep{ravi2024sam2} and HQ-SAM~2~\citep{hqsam2}, use their own vision encoder and do not use Moondream vision features, while our refiner conditions on Moondream vision tokens.

\begin{table}[t]
  \centering
  \caption{Refiner comparison. mIoU (\%) on a held-out set of RefCOCO-M rollout-derived coarse masks conditioned on the ground-truth bounding box (higher is better).}
  \label{tab:ablations}
  \begin{tabular}{lc}
    \toprule
    Refiner & RefCOCO-M\\
    \midrule
    SAM 2 & 90.8\\
    HQ-SAM 2 & 91.7\\
    Moondream Seg. & \textbf{93.8}\\
    \bottomrule
  \end{tabular}
\end{table}

To select the number of refinement steps, we train separate refiners with $T$ ranging from 1 to 10 and evaluate each at its training step count (Figure~\ref{fig:refinement_steps}). $T=5$ is selected as it is approximately the point of diminishing returns.

\section{Limitations and Future Work}
Our vector-path representation can struggle with disjoint regions and extremely thin structures. Multi-instance masks require multiple passes: a first pass to obtain a point or bounding box for each instance, followed by a second pass for each instance conditioned on the text prompt and spatial reference. Future work includes direct multi-instance masking and extensions to video.

\section{Conclusion}
We presented Moondream Segmentation, a VLM-based referring segmentation model that combines compact vector-path prediction with iterative mask refinement. Across RefCOCO-family benchmarks, it achieves state-of-the-art results on multiple splits and ties SAM~3 at 62.6 mIoU on LVIS. We introduce RefCOCO-M to reduce evaluation noise from coarse polygon annotations and better measure boundary quality. Together, these results show that compact vector intermediates paired with iterative refinement are an effective way to bring high-fidelity segmentation to VLMs.

\section{Impact Statement}
Moondream Segmentation enables language-driven, pixel-accurate mask generation with sharp boundaries, which can improve workflows such as image editing, compositing, and data annotation. The accompanying RefCOCO-M masks can reduce evaluation noise and support more reliable benchmarking of boundary quality. Like other segmentation models, it could also be misused for non-consensual image manipulation or creating misleading media.

\section{Acknowledgments}
We thank Vik Korrapati and Jay Allen for their guidance and contributions on this work.

\bibliographystyle{icml2026/icml2026}
\balance
\bibliography{refs}

\clearpage
\appendix
\onecolumn
\raggedbottom
\section{Appendix}
\subsection{Vector Path Tokenization and Rasterization}
\label{app:svg}
\paragraph{Segmentation prompt and answer sequence.}
A single segmentation query is serialized as one token sequence. We use a PrefixLM-style layout where the image token grid is inserted immediately after the first token (the BOS token). Conceptually, the sequence is:
\begin{center}
\fbox{\begin{minipage}{0.97\columnwidth}
\footnotesize
\begin{tabular}{@{}p{0.93\columnwidth}@{}}
\texttt{<BOS> [IMG] x 729}\\
\texttt{<|mode|> segment <|prompt|> [optional spatial prompt] <referring expression>}\\
\texttt{<|answer|> <|coord|> <|coord|> <|size|> [<|svg\_start|>] <vector path tokens> [<|svg\_end|>] <EOS>}
\end{tabular}
\end{minipage}}
\end{center}
The optional spatial prompt can be a point prompt (\texttt{<|coord|><|coord|>}) or a box prompt (\texttt{<|coord|><|coord|><|size|>}) placed after \texttt{<|prompt|>} and before the referring expression. The answer begins with a predicted box (center and size) represented by \texttt{<|coord|><|coord|><|size|>} and decoded by the RegionModel (Appendix~\ref{app:regionmodel}).

\paragraph{Vector path token grammar.}
In the vector path portion of the answer (optionally bracketed by \texttt{<|svg\_start|>} and \texttt{<|svg\_end|>}), the model emits a flat stream of vector path tokens:
\begin{itemize}
  \item \textbf{Commands:} \texttt{M}, \texttt{L}, \texttt{C}, \texttt{Z}/\texttt{z} (absolute commands only).
  \item \textbf{Integers:} coordinate values are quantized to integers, each represented by a single text token (we use values in $[-39,999]$ and typically scale a $[0,1]$ box-normalized path by a viewBox size of $960$).
  \item \textbf{Negative values:} a negative integer is represented as the two-token sequence \texttt{-} followed by its magnitude. The token \texttt{-} only has meaning as a sign for the immediately following integer token.
\end{itemize}
Parsing is deterministic: we read a command token, then consume all following integer tokens until the next command token. Integers are grouped into $(x,y)$ pairs. For a \texttt{C} command, the sequence is interpreted as triples of pairs (control point 1, control point 2, end point). We do not require the model to emit separators such as commas; instead, separators are inserted deterministically when converting the token stream into an SVG \texttt{d} string.

\paragraph{Rasterization.}
Given a predicted box $b=(c_x,c_y,w,h)$ in normalized image coordinates and a decoded path $p$, we serialize the vector path as an SVG string with a fixed viewBox and use an affine \texttt{translate+scale} to map the box-normalized path into the image. We rasterize the resulting SVG deterministically to obtain $\tilde{M}^{(0)}_{\text{nat}}$ at $(H,W)$, then resize to the refiner working resolution $(H_0,W_0)$.

\subsection{RegionModel Encoding and Decoding}
\label{app:regionmodel}
\paragraph{Overview.}
Moondream represents binned spatial values using special tokens \texttt{<|coord|>} and \texttt{<|size|>}. A point is represented as \texttt{<|coord|><|coord|>} (center $(c_x,c_y)$). A box is represented as \texttt{<|coord|><|coord|><|size|>} (center $(c_x,c_y)$ and size $(w,h)$). When a spatial token is generated, the RegionModel decodes a discretized value from the current hidden state and we immediately re-embed the resulting real value for the next decoding step (``decode-then-re-embed'').

\paragraph{Fourier features and encoders.}
We use fixed random Fourier features~\citep{tancik2020fourier}. For an input $u\in\mathbb{R}^{d_{\text{in}}}$ and a random matrix $W\in\mathbb{R}^{(d_{\phi}/2)\times d_{\text{in}}}$, we define
\begin{equation}
\phi(u)=\big[\cos(2\pi uW^\top),\;\sin(2\pi uW^\top)\big]\in\mathbb{R}^{d_{\phi}}.
\end{equation}
Let $D$ be the transformer width (e.g., $D=2048$). The RegionModel maps a scalar coordinate $c\in[0,1]$ to an embedding $\mathbf{e}_{\text{coord}}(c)\in\mathbb{R}^{D}$, and maps a size vector $\mathbf{s}=(w,h)\in(0,1]^2$ to an embedding $\mathbf{e}_{\text{size}}(\mathbf{s})\in\mathbb{R}^{D}$ using learned linear projections of Fourier features.

\paragraph{Decoders.}
Let $\mathrm{LN}(\cdot)$ denote layer normalization and let $V=1024$ be the number of bins. Given a hidden state $\mathbf{h}\in\mathbb{R}^{D}$ at a spatial token position, we compute
\begin{equation}
\tilde{\mathbf{h}}=\mathrm{LN}(\mathbf{h}),
\end{equation}
and decode logits with linear maps:
\begin{equation}
\begin{aligned}
\mathbf{z}_{\text{coord}} &= W_{\text{coord}}\tilde{\mathbf{h}}+\mathbf{b}_{\text{coord}} \in\mathbb{R}^{V},\\
\mathbf{z}_{\text{size}} &= \mathrm{reshape}\!\Big(W_{\text{size}}\tilde{\mathbf{h}}+\mathbf{b}_{\text{size}},\,2,\,V\Big) \in\mathbb{R}^{2\times V},
\end{aligned}
\end{equation}
where $W_{\text{coord}}\in\mathbb{R}^{V\times D}$, $\mathbf{b}_{\text{coord}}\in\mathbb{R}^{V}$, $W_{\text{size}}\in\mathbb{R}^{(2V)\times D}$, and $\mathbf{b}_{\text{size}}\in\mathbb{R}^{2V}$. The two rows of $\mathbf{z}_{\text{size}}$ correspond to width and height logits respectively.

\paragraph{De-quantization for decode-then-re-embed.}
We interpret a coordinate bin $v\in\{0,\dots,V-1\}$ as the real value
\begin{equation}
\hat{c}=\frac{v}{V}.
\end{equation}
For sizes we use log-space bins. Given a size bin $v\in\{0,\dots,V-1\}$, we map it to a positive extent
\begin{equation}
\hat{s}=2^{\left(\frac{v}{V-1}\right)\cdot 10 - 10},
\end{equation}
and apply this independently to $(v_w,v_h)$ to obtain $(\hat{w},\hat{h})$.

\begin{algorithm}[t]
\caption{RegionModel decode-then-re-embed}
\label{alg:regionmodel}
\begin{algorithmic}[1]
\STATE \textbf{Input:} next token ids $y_t$; hidden states $\mathbf{h}_t$; embeddings $\mathbf{e}_{\text{coord}}(\cdot),\mathbf{e}_{\text{size}}(\cdot)$; bins $V=1024$
\FOR{each position $i$ where $y_t[i]=\texttt{<|coord|>}$}
  \STATE $\tilde{\mathbf{h}}\leftarrow \mathrm{LN}(\mathbf{h}_t[i])$
  \STATE $\mathbf{z}\leftarrow W_{\text{coord}}\tilde{\mathbf{h}}+\mathbf{b}_{\text{coord}} \in \mathbb{R}^{V}$
  \STATE $v \leftarrow \arg\max \mathbf{z}$ \COMMENT{or sample from $\mathrm{softmax}(\mathbf{z})$}
  \STATE $\hat{c} \leftarrow v / V$
  \STATE Replace the transformer input embedding at position $i$ with $\mathbf{e}_{\text{coord}}(\hat{c})$
\ENDFOR
\FOR{each position $i$ where $y_t[i]=\texttt{<|size|>}$}
  \STATE $\tilde{\mathbf{h}}\leftarrow \mathrm{LN}(\mathbf{h}_t[i])$
  \STATE $\mathbf{z}\leftarrow \mathrm{reshape}\!\Big(W_{\text{size}}\tilde{\mathbf{h}}+\mathbf{b}_{\text{size}},\,2,\,V\Big) \in \mathbb{R}^{2\times V}$
  \STATE $(v_w,v_h)\leftarrow \arg\max \mathbf{z}$ \COMMENT{per row (or sample)}
  \STATE $(\hat{w},\hat{h}) \leftarrow \left(2^{(v_w/(V-1))\cdot 10 - 10},\;2^{(v_h/(V-1))\cdot 10 - 10}\right)$
  \STATE Replace the transformer input embedding at position $i$ with $\mathbf{e}_{\text{size}}((\hat{w},\hat{h}))$
\ENDFOR
\end{algorithmic}
\end{algorithm}

\subsection{Refiner Inference Loop}
\label{app:refiner_inference}
For completeness, we provide the refiner inference procedure used at test time. Given frozen encoder features $(F_{\text{early}},F_{\text{final}})$ and an initial coarse mask $\tilde{M}^{(0)}$ (obtained by rasterizing the decoded vector path), we iteratively refine for $T$ steps by selecting the highest-scoring mask at each iteration.

\begin{algorithm}[t]
\caption{Refiner inference (iterative mask refinement)}
\label{alg:refiner_inference}
\begin{algorithmic}[1]
\STATE \textbf{Input:} frozen features $F_{\text{final}},F_{\text{early}}$; initial mask $\tilde{M}^{(0)}$; steps $T$; refiner $f_{\phi}$
\STATE $\tilde{M}\leftarrow \tilde{M}^{(0)}$
\FOR{$t=0$ to $T-1$}
  \STATE $(L^{(t)}, q^{(t)}) \leftarrow f_{\phi}(F_{\text{final}},F_{\text{early}},\tilde{M})$
  \STATE $m^* \leftarrow \arg\max_m q^{(t)}_m$
  \STATE $\tilde{M} \leftarrow \sigma(L^{(t)}_{m^*})$
\ENDFOR
\STATE \textbf{Return:} $\tilde{M}$
\end{algorithmic}
\end{algorithm}

\subsection{Implementation Details}
\label{app:impl}
\paragraph{Refiner training schedule.}
We train the refiner by unrolling $T=5$ refinement iterations at a fixed refiner resolution ($H_0=W_0=378$). The vision encoder is frozen and the refiner is optimized with AdamW.

\paragraph{Boundary-weighted loss and warmup.}
To emphasize edge alignment, we add a boundary-weighted BCE term based on a signed distance transform of the target mask. Let $D_{\text{dist}}\in\mathbb{R}^{H_0\times W_0}$ be a normalized signed distance map (negative inside the mask, positive outside, and $D_{\text{dist}}=0$ on the boundary), and define per-pixel weights $w_{ij}=\exp(-\gamma|D_{\text{dist}}(i,j)|)$ (we use $\gamma=10$). For predicted probabilities $P\in[0,1]^{H_0\times W_0}$ and target $M\in\{0,1\}^{H_0\times W_0}$, we use
\begin{equation}
\mathrm{BoundaryLoss}(P,M)=\frac{\sum_{i,j} w_{ij}\,\mathrm{BCE}\big(P_{ij},M_{ij}\big)}{\sum_{i,j} w_{ij}+\varepsilon}.
\end{equation}
We apply a warmup schedule on its weight $\lambda_{\partial}(s)$ over global training step $s$: $\lambda_{\partial}(s)=0$ for the first $1000$ steps, then it increases linearly to $1$ over the next $500$ steps and remains at $1$ thereafter.

\paragraph{SoftIoU.}
\label{app:softiou}
For predicted probabilities $P\in[0,1]^{H_0\times W_0}$ and target $M\in\{0,1\}^{H_0\times W_0}$, we define
\begin{equation}
\mathrm{SoftIoU}(P,M)=\frac{\sum_{i,j} P_{ij}\,M_{ij}}{\sum_{i,j} P_{ij}+\sum_{i,j} M_{ij}-\sum_{i,j} P_{ij}\,M_{ij}}.
\end{equation}

\subsection{RefCOCO-M Additional Details}

\begin{figure}[H]
  \centering
  \includegraphics[width=0.72\linewidth]{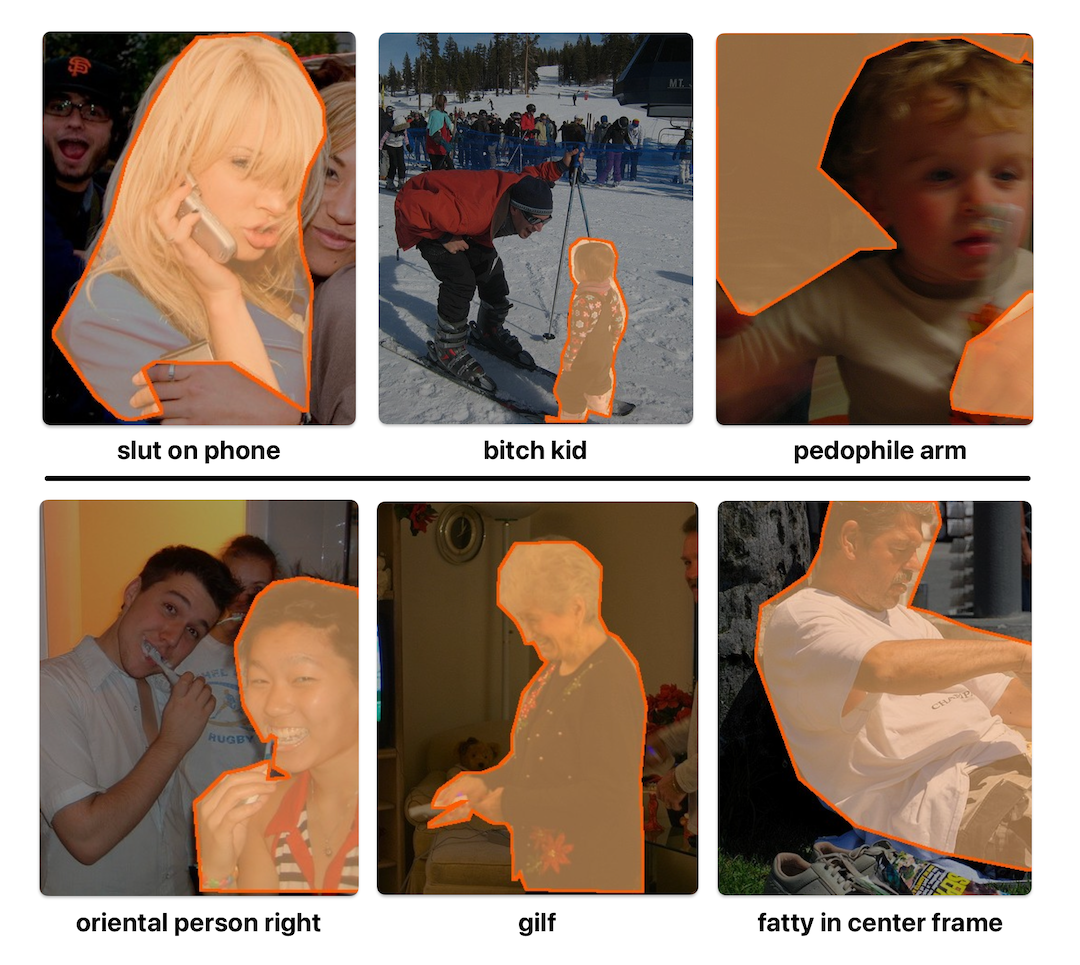}
  \caption{Examples of referring expressions removed from RefCOCO-M by the safety pipeline.}
  \label{fig:filtered_samples}
\end{figure}

\subsection{Additional Qualitative Samples}

\begin{figure}[H]
  \centering
  \includegraphics[width=\linewidth,height=0.82\textheight,keepaspectratio]{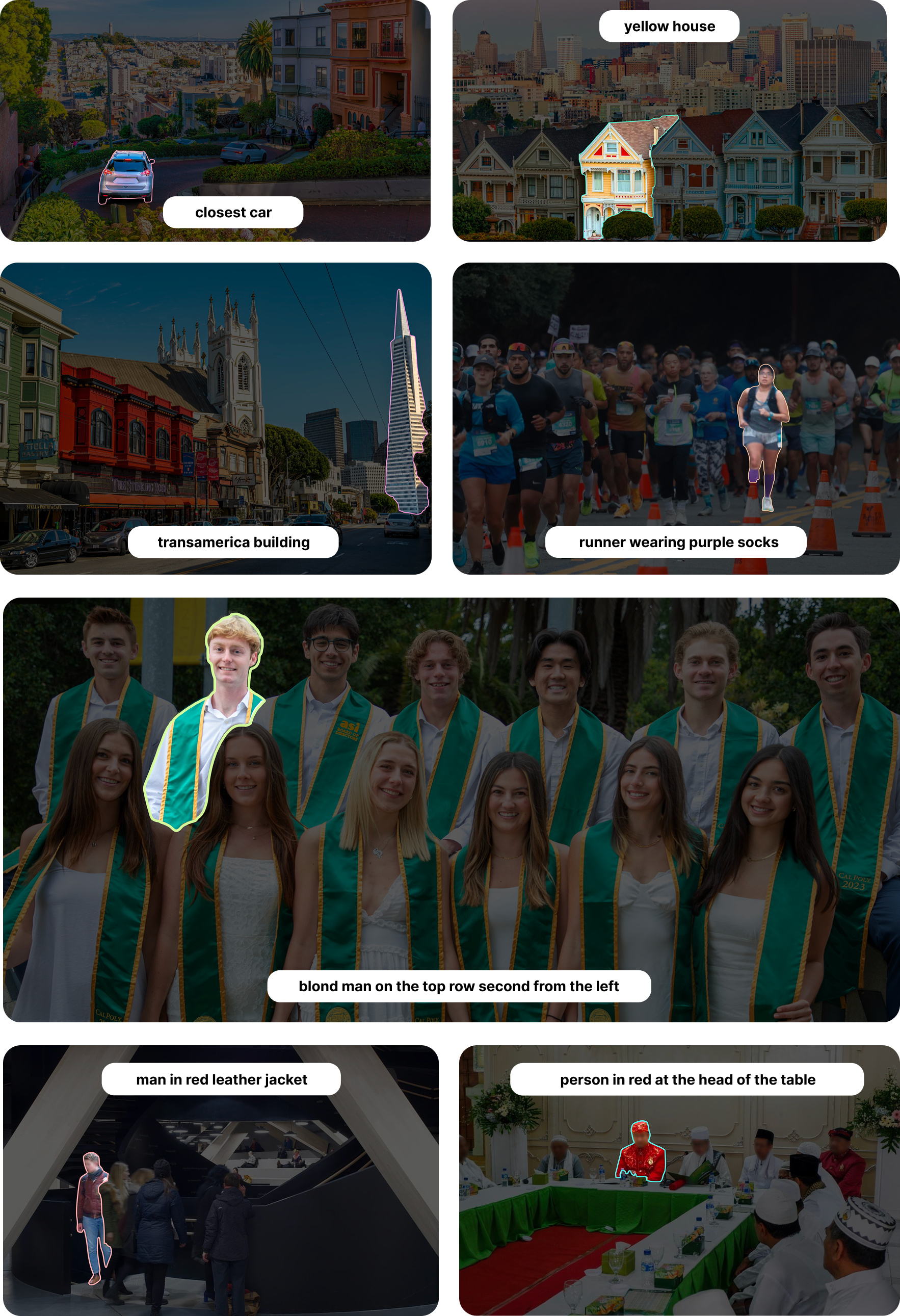}
  \caption{Additional qualitative samples produced by Moondream Segmentation. Prompts are shown in white boxes.}
  \label{fig:additional_qual}
\end{figure}

\begin{figure}[H]
  \centering
  \includegraphics[width=\linewidth,height=0.82\textheight,keepaspectratio]{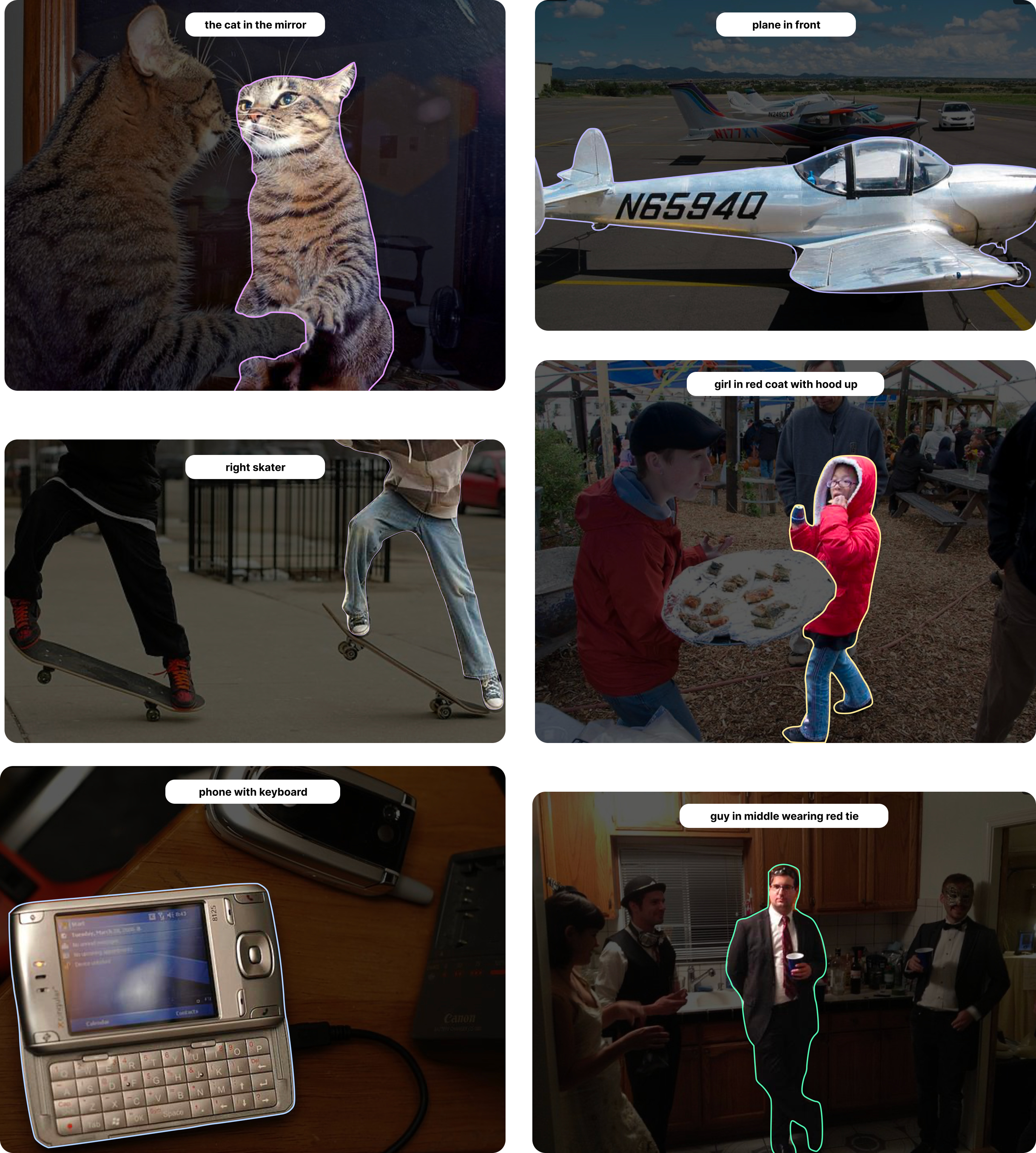}
  \caption{RefCOCO validation samples produced by Moondream Segmentation. Prompts are shown in white boxes.}
  \label{fig:refcoco_qual}
\end{figure}

\begin{figure}[H]
  \centering
  \includegraphics[width=\linewidth,height=0.82\textheight,keepaspectratio]{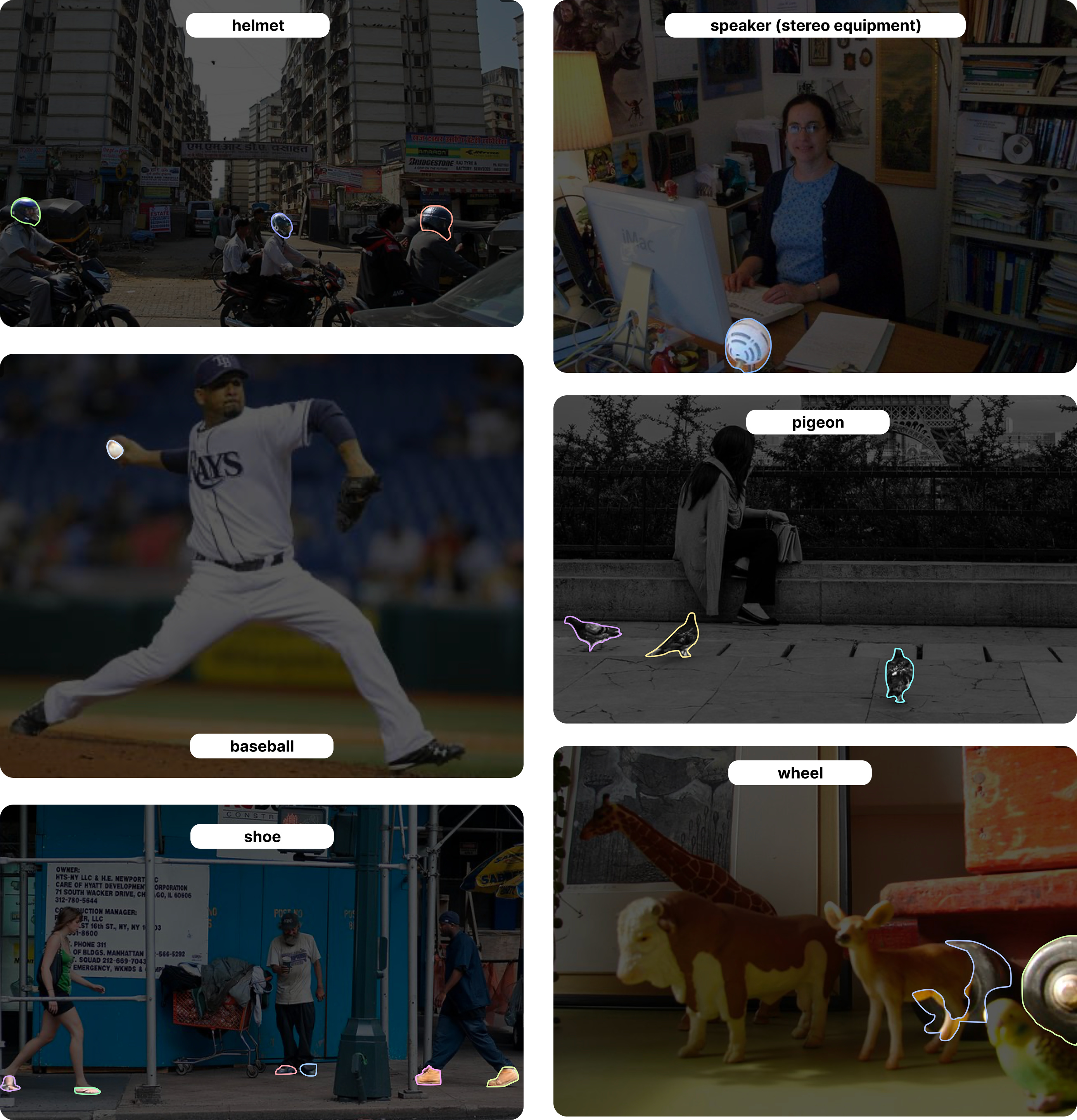}
  \caption{LVIS validation samples produced by Moondream Segmentation. Prompts are shown in white boxes.}
  \label{fig:lvis_qual}
\end{figure}

\end{document}